\documentclass[sigconf,screen,nonacm]{acmart}
\renewcommand\footnotetextcopyrightpermission[1]{}
\AtBeginDocument{%
  }

\usepackage{booktabs}
\usepackage{multirow}
\usepackage{graphicx}
\usepackage[table]{xcolor}

\begin{document}

\title{Direct Discrepancy Replay: Distribution-Discrepancy Condensation and Manifold-Consistent Replay for Continual Face Forgery Detection}

\author{Tianshuo Zhang\textsuperscript{\rm 1,2${\dag}$},Haoyuan Zhang\textsuperscript{\rm 1,2${\dag}$},Siran Peng\textsuperscript{\rm 2,1},Weisong Zhao\textsuperscript{\rm 3,4},Xiangyu Zhu\textsuperscript{\rm 1,2},Zhen Lei\textsuperscript{\rm 1,2,5}}
\authornote{Corresponding author. \hspace{2mm} ${\dag}$Equal Contribution.}
\affiliation{%
	\institution{
    $^{1}$SAI, UCAS,\ \ \ 
    $^{2}$MAIS, CASIA,\ \ \  
    $^{3}$IIE, CAS,\ \ \ 
    $^{4}$SCS, UCAS,\ \ \ 
    $^{5}$SCSE, FIE, M.U.S.T\\
    {\small \tt tianshuo.zhang@nlpr.ia.ac.cn, \{zhanghaoyuan2023, pengsiran2023, xiangyu.zhu, zhen.lei\}@ia.ac.cn}
    }
    \city{}
	\country{}
}

\renewcommand{\shortauthors}{Zhang et al.}

\begin{abstract}
Continual face forgery detection (CFFD) requires detectors to learn emerging forgery paradigms without forgetting previously seen manipulations. Existing CFFD methods commonly rely on replaying a small amount of past data to mitigate forgetting. Such replay is typically implemented either by storing a few historical samples or by synthesizing pseudo-forgeries from detector-dependent perturbations. Under strict memory budgets, the former cannot adequately cover diverse forgery cues and may expose facial identities, while the latter remains strongly tied to past decision boundaries. We argue that the core role of replay in CFFD is to reinstate the distributions of previous forgery tasks during subsequent training. To this end, we directly condense the discrepancy between real and fake distributions and leverage real faces from the current stage to perform distribution-level replay. Specifically, we introduce Distribution-Discrepancy Condensation (DDC), which models the real-to-fake discrepancy via a surrogate factorization in characteristic-function space and condenses it into a tiny bank of distribution discrepancy maps. We further propose Manifold-Consistent Replay (MCR), which synthesizes replay samples through variance-preserving composition of these maps with current-stage real faces, yielding samples that reflect previous-task forgery cues while remaining compatible with current real-face statistics. Operating under an extremely small memory budget and without directly storing raw historical face images, our framework consistently outperforms prior CFFD baselines and significantly mitigates catastrophic forgetting. Replay-level privacy analysis further suggests reduced identity leakage risk relative to selection-based replay.
\end{abstract}





\maketitle

\section{Introduction}
\begin{figure}[t]
    \centering
    \includegraphics[width=\linewidth]{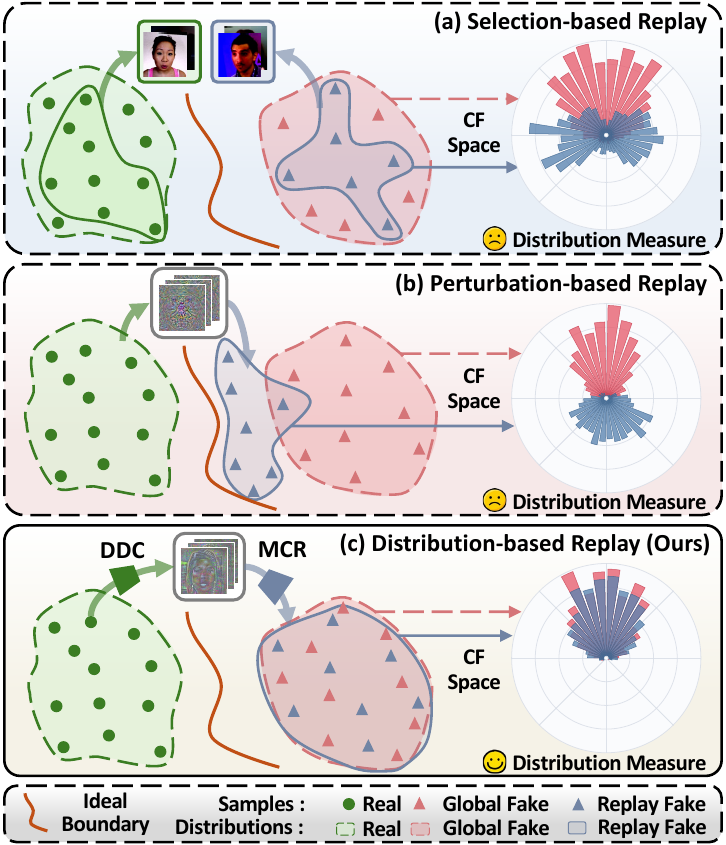}
    \caption{Comparison of replay paradigms in CFFD. Selection-based replay uses sparse support samples and only coarsely approximates the previous fake distribution. Perturbation-based replay preserves boundary-related signals but ignores distribution alignment. Our method condenses the real-to-fake discrepancy and unfolds it through MCR in the next stage to produce replay samples that better match the previous fake distribution. Radial charts show a CF-based distribution measure; closer overlap means better replay fidelity.}
    \label{fig:first}
\end{figure}

Face generation and manipulation techniques built on GANs~\cite{gan_gen2,gan_gen3} and diffusion~\cite{diff2,diff3} models are evolving rapidly, with forgery families emerging sequentially. In real deployments, repeatedly retaining historical face data and re-training detectors from scratch is impractical due to privacy, storage, and computational constraints. These challenges make face forgery detection a naturally continual problem, where detectors~\cite{sbi,moeffd} must absorb emerging forgeries through incremental updates while retaining their capability to recognize manipulations from earlier tasks. In practice, however, such sequential adaptation often erodes the model’s sensitivity to learned forgery cues, leading to severe catastrophic forgetting. Replay~\cite{replay,er,gem} thus becomes a natural solution for continual face forgery detection (CFFD). Importantly, replay should not be viewed as merely rehearsing a handful of stored samples. Its fundamental role is to continually provide training signals that preserve the forgery statistics of prior tasks during later-stage learning.

As shown in Fig.~\ref{fig:first}, existing replay methods for CFFD can be broadly divided into two categories. The first~\cite{dfil,surlid} retains a compact subset of historical samples and replays them during later training. Under strict memory budgets, however, such methods often fail to cover the diversity of forgery cues from previous tasks and may raise privacy concerns due to identity retention. The second category~\cite{uap,hdp} restores knowledge through perturbations. While effective to some extent, such replay is still tied to past decision boundaries and is therefore less robust to backbone updates and representation drift. Overall, these methods only approximate replay indirectly, which motivates a more direct formulation.

\begin{figure*}[t]
    \centering
    \includegraphics[width=\linewidth]{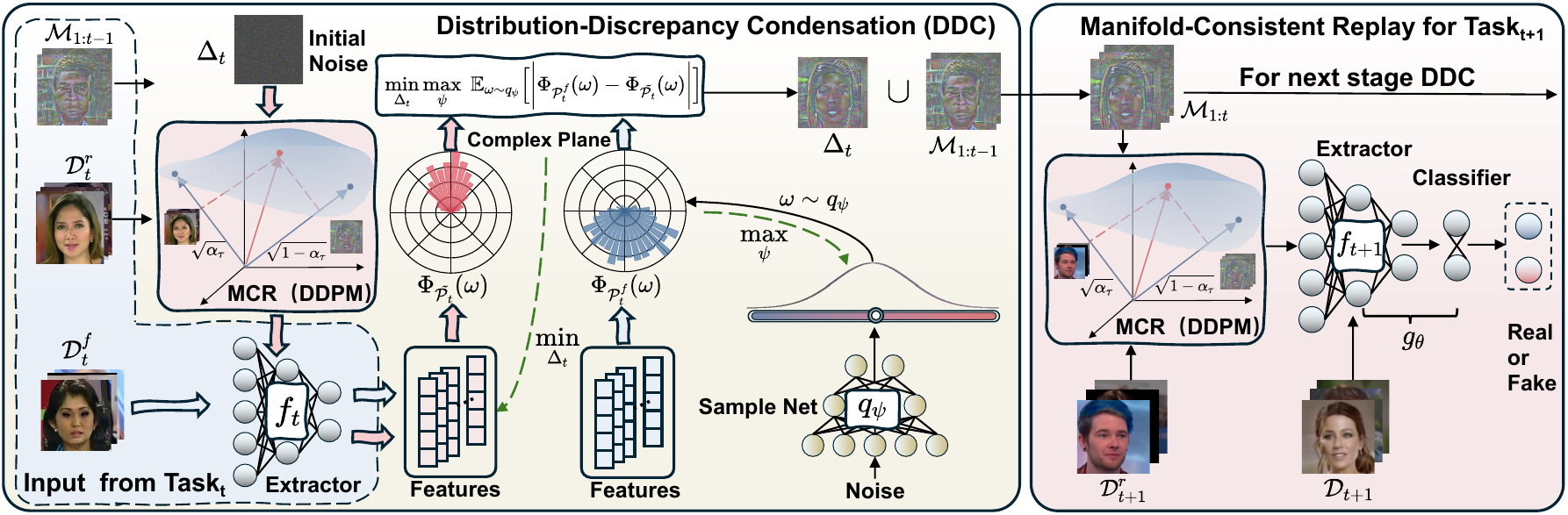}
    \caption{Our method models the distribution discrepancy between real and fake samples via a surrogate factorization in characteristic-function (CF) space, and transforms it into compact discrepancy maps (DDMs) that can be composed with real faces to synthesize fake samples. Distribution-Discrepancy Condensation (DDC) adversarially optimizes these maps by matching the CFs of target fake samples and synthesized samples: a sample network learns the frequency distribution $q_{\psi}(\omega)$ that maximizes their discrepancy, while the learnable DDMs minimize it under the sampled $\omega$. To ensure that synthesized samples remain compatible with natural images for joint training, Manifold-Consistent Replay (MCR) achieves the energy manifold requirement with a variance-preserving constraint, which naturally yields a DDPM-style composition. At task $t{+}1$, the stored DDMs are unfolded on current real faces to generate replay fake samples that match previous forgery distributions.}
    \label{fig:mian}
\end{figure*}

To address this issue, as shown in Fig.~\ref{fig:mian}, we propose directly replaying the real-to-fake distribution discrepancy of previous tasks in CFFD. By introducing characteristic functions (CFs)~\cite{cf} as the distribution metric, we use a surrogate factorization in CF space to describe this discrepancy and translate it into an optimizable spatial-domain formulation, which can be instantiated as a tiny bank of distribution discrepancy maps (DDMs) for replay. This gives rise to our Distribution-Discrepancy Condensation (DDC), which provides a compact discrepancy memory without directly storing raw historical face images. When composed with abundant real faces in later stages, these DDMs yield diverse replay samples that remain statistically aligned with previous forgery distributions.

Obtaining compact discrepancy memories alone is insufficient for continual learning. In later stages, replay samples must be jointly trained with current-task data, requiring the samples synthesized from DDMs to remain on the same energy manifold as current real faces while staying statistically aligned with previous-task forgery distributions. In practice, we achieve this requirement with an implementable variance-preserving constraint on the composition of DDMs with current real faces, rather than enforcing a strict manifold correspondence directly, which naturally yields a DDPM-style~\cite{ddpm} formulation. Based on this insight, we propose Manifold-Consistent Replay (MCR), which synthesizes replay fake samples for joint training under this formulation.

By storing only lightweight DDMs instead of historical face images, our method avoids directly storing raw historical face images and yields substantially more diverse replay samples by composing DDMs with abundant real faces from the subsequent stage. Extensive experiments show that, under an extremely small memory budget, our method not only significantly mitigates catastrophic forgetting, but also consistently outperforms existing CFFD methods, including several methods that use substantially larger memory. Our contributions are summarized as follows.

\begin{itemize}
\item We directly model replay in CFFD as the real-to-fake distribution discrepancy of previous tasks, rather than relying on retained samples or boundary-tied perturbations.

\item We propose Distribution-Discrepancy Condensation (DDC), which formulates the discrepancy via a surrogate factorization in characteristic-function space and instantiates it as a tiny bank of optimizable distribution discrepancy maps.

\item We propose Manifold-Consistent Replay (MCR), which aims to keep replay samples on the same energy manifold as current real faces and achieves this through a DDPM-style variance-preserving constraint for unfolding stored DDMs on abundant current real faces.
\end{itemize}

\section{Related Works}

\subsection{Dataset Condensation}

Dataset condensation~\cite{dd} aims to compress a large dataset into a small synthetic set such that models trained on it match the performance of full-dataset training.
Early methods frame this as a bilevel meta-learning problem~\cite{dd}, which is computationally prohibitive.
Gradient matching~\cite{dc,dsa} alleviates this by aligning per-step gradients between real and synthetic data, while trajectory matching~\cite{mtt,datm} extends the alignment to multi-step training trajectories, achieving near-lossless compression on small benchmarks.
Distribution matching~\cite{dm,cafe,idm,m3d} offers a more efficient alternative: rather than unrolling optimization, it matches feature statistics or distributions between real and synthetic data in embedding spaces, typically via mean matching~\cite{dm}, layer-wise feature alignment~\cite{cafe}, or Maximum Mean Discrepancy~\cite{m3d}.
A concurrent advance, NCFM~\cite{ncfm}, reformulates distribution matching through characteristic-function (CF) discrepancy and employs a min--max scheme to adaptively emphasize informative frequency components. 
In this work, we build on this CF-based matching perspective, but repurpose them to condense the lightweight \emph{real-to-fake discrepancy} specific to each forgery task.

\subsection{Continual Face Forgery Detection}

Continual face forgery detection (CFFD) requires a detector to sequentially learn emerging forgery types without forgetting previously seen ones.
General continual learning methods address catastrophic forgetting through regularization~\cite{ewc,lwf,si}, which constrains parameter updates to preserve past knowledge, or replay~\cite{er,icarl,dgr}, which rehearses stored or generated samples from previous tasks.
Domain-specific CFFD methods build on these paradigms.
CoReD~\cite{cored} applies knowledge distillation to transfer generalizable forgery representations across tasks.
DFIL~\cite{dfil} exploits domain-invariant forgery clues to reduce cross-task interference.
DMP~\cite{dmp} maintains dynamic mixed prototypes to represent distributions of past forgery tasks without storing raw samples.
HDP~\cite{hdp} preserves historical distributions by simulating past forgery patterns with universal adversarial perturbations and applying knowledge distillation to constrain real-face distribution drift.
SUR-LID~\cite{surlid} proposes sparse uniform replay combined with latent-space feature alignment to isolate and preserve task-specific feature distributions.
To overcome the privacy risks of storing raw faces and the backbone-fragility of perturbations in existing replay-based CFFD methods, we retain only compact distribution discrepancy maps that encode real-to-fake cues without exposing raw faces.

\section{Methodology}

\subsection{Problem Setup and Overview}
We consider continual face forgery detection (CFFD) over a sequence of tasks
$\{\mathcal{T}_t\}_{t=1}^{T}$. Each task $\mathcal{T}_t$ provides a binary forgery
detection dataset
\begin{equation}
\mathcal{D}_t = \mathcal{D}_t^{r} \cup \mathcal{D}_t^{f},
\end{equation}
where $\mathcal{D}_t^{r}=\{x_t^{r}\}$ denotes the set of real faces sampled from the real-face distribution $\mathcal{P}_t^{r}$, and $\mathcal{D}_t^{f}=\{x_t^{f}\}$ denotes the set of forged faces sampled from the fake-face distribution $\mathcal{P}_t^{f}$. The detector is denoted by $g_{\theta}$, and its penultimate feature extractor is denoted by $f(\cdot)$.
At task $t$, the model has access only to the current dataset $\mathcal{D}_t$ and a
compact memory from previous tasks; all raw images from earlier tasks are assumed to
be unavailable.

Unlike conventional replay methods that retain historical samples or store
detector-dependent perturbation surrogates, our method stores only a tiny set of
\emph{distribution discrepancy maps} (DDMs). Specifically, after finishing task $t$,
we preserve a compact bank
\begin{equation}
\Delta_t=\{d_{t,k}\}_{k=1}^{K},
\end{equation}
where each $d_{t,k}\in\mathbb{R}^{C\times H\times W}$ is a learnable discrepancy map
that encodes the real-to-fake distribution discrepancy of task $t$. The overall memory
after task $t-1$ is
\begin{equation}
\mathcal{M}_{1:t-1}=\{\Delta_s\}_{s=1}^{t-1}.
\end{equation}

As shown in Fig.~\ref{fig:mian}, our framework consists of two components. First, \emph{Distribution-Discrepancy Condensation} (DDC) compresses the real-to-fake distribution discrepancy of the completed task into a tiny DDM bank, which serves as an empirical surrogate of the task-specific discrepancy. Second, \emph{Manifold-Consistent Replay} (MCR) unfolds the stored DDMs by composing them with abundant real faces from the next stage, thereby generating replay fake samples for joint training without directly storing raw historical face images.

\subsection{Distribution-Discrepancy Condensation}

\noindent{\textbf{\textit{Surrogate Discrepancy Formulation via CFs.}}} In CFFD, real and fake faces share highly similar semantic content, while their discriminative differences are mainly reflected in subtle forgery traces. This suggests that replay should target the discrepancy between real and fake distributions, rather than the images themselves. To characterize such discrepancy, we introduce \emph{characteristic functions} (CFs) as the distribution metric. For a distribution $\mathcal{P}$ over images, we define its feature-space CF as
\begin{equation}
\Phi_{\mathcal{P}}(\omega)
=
\mathbb{E}_{x\sim\mathcal{P}}
\left[
\exp\!\left(j \left\langle \omega, f(x) \right\rangle\right)
\right],
\end{equation}
where $\omega$ is the frequency variable and $j=\sqrt{-1}$ is the imaginary unit. We introduce a surrogate discrepancy distribution $\mathcal{P}_t^{\Delta}$ and write the real-to-fake discrepancy of task $t$ as a multiplicative factor in CF space:
\begin{equation}
\Phi_{\mathcal{P}_t^{f}}(\omega)
\approx
\Phi_{\mathcal{P}_t^{r}}(a\omega)\,
\Phi_{\mathcal{P}_t^{\Delta}}(b\omega),
\label{eq:cf_factorization}
\end{equation}
where $a,b>0$ are composition coefficients. Eq.~\eqref{eq:cf_factorization} is used as a \emph{surrogate factorization}: in general, the ratio $\Phi_{\mathcal{P}_t^{f}}(\omega)/\Phi_{\mathcal{P}_t^{r}}(a\omega)$ need not define a valid characteristic function, and the denominator may vanish. Accordingly, $\mathcal{P}_t^{\Delta}$ should be understood as a compact surrogate object rather than a rigorously identified distributional factor. Our method does not rely on the exact validity of Eq.~\eqref{eq:cf_factorization}; instead, it is driven by the empirical CF matching objective introduced below.

\noindent{\textbf{\textit{From CF-space Discrepancy to Spatial-Domain DDMs.}}} Directly optimizing the multiplicative discrepancy in Eq.~\eqref{eq:cf_factorization} is challenging, since it is defined over complex-valued CFs and requires differentiation through exponential and multiplicative terms. We therefore move from the CF-space formulation to an image-space realization, where DDMs are learned at the input resolution. Strictly speaking, the additive realization below is derived for pixel-space random variables, whereas the CF matching objective~\cite{ncfm} is evaluated in the frozen feature space defined by $f(\cdot)$ during discrepancy condensation. Although $f(\cdot)$ is nonlinear, linear compositions in input or representation space are widely used as vicinal relaxations~\cite{mixup,m_mixup}; here, variance-preserving pixel-space blending defines a smooth local noise family, while the end-to-end CF matching objective encourages synthesized samples to align with target fake samples.

Let $X\sim\mathcal{P}_t^{r}$ denote a real-face random variable and
$\Delta\sim\mathcal{P}_t^{\Delta}$ denote a discrepancy random variable, with
$X\perp\Delta$. By the scaling property of characteristic functions, for any random
variable $Z$ and scalar $c$, we have
\begin{equation}
\Phi_{cZ}(\omega)=\Phi_Z(c\omega).
\label{eq:cf_scaling}
\end{equation}
Applying Eq.~\eqref{eq:cf_scaling} to Eq.~\eqref{eq:cf_factorization} gives
\begin{equation}
\Phi_{\mathcal{P}_t^{f}}(\omega)
\approx
\Phi_X(a\omega)\,\Phi_{\Delta}(b\omega)
=
\Phi_{aX}(\omega)\,\Phi_{b\Delta}(\omega).
\label{eq:cf_factorization_scaled}
\end{equation}

Since $X\perp\Delta$, the scaled variables $aX$ and $b\Delta$ remain independent. By
the convolution theorem of probability measures, the characteristic function of the sum
of two independent random variables equals the product of their characteristic
functions. Therefore,
\begin{equation}
\Phi_{aX+b\Delta}(\omega)
=
\Phi_{aX}(\omega)\,\Phi_{b\Delta}(\omega).
\label{eq:cf_spatial_equivalence}
\end{equation}
Combining Eq.~\eqref{eq:cf_factorization_scaled} and
Eq.~\eqref{eq:cf_spatial_equivalence} yields
\begin{equation}
\Phi_{\mathcal{P}_t^{f}}(\omega)
\approx
\Phi_{aX+b\Delta}(\omega),
\label{eq:cf_to_spatial}
\end{equation}
which means that the multiplicative discrepancy in CF space admits an additive
realization in the spatial domain. Equivalently, the fake distribution can be
approximated by the distribution of the composed variable $aX+b\Delta$. Accordingly, we
instantiate replay samples as
\begin{equation}
\tilde{x}_t
=
a x_t^{r} + b \delta_t,
\qquad
x_t^{r}\sim\mathcal{P}_t^{r},\;
\delta_t\sim\mathcal{P}_t^{\Delta}.
\label{eq:spatial_composition}
\end{equation}

This observation motivates the need for a compact set of spatial-domain discrepancy elements that instantiate this surrogate discrepancy, rather than learning an explicit generator or storing historical images. We instantiate this surrogate discrepancy with a tiny bank of distribution discrepancy maps (DDMs),
\begin{equation}
\Delta_t=\{d_{t,k}\}_{k=1}^{K},
\end{equation}
where each $d_{t,k}\in\mathbb{R}^{C\times H\times W}$ serves as a support element of the spatial-domain surrogate discrepancy.

\noindent{\textbf{\textit{DDC Objective and Memory Update.}}} Given a real batch $B_t^{r}\subset\mathcal{D}_t^{r}$ and a fake batch
$B_t^{f}\subset\mathcal{D}_t^{f}$ from task $t$, DDC learns the DDM bank
$\Delta_t=\{d_{t,k}\}_{k=1}^{K}$ by matching the fake distribution with the
distribution induced by composing real faces and discrepancy maps. Following
Eq.~\eqref{eq:spatial_composition}, we construct a synthetic batch
$\tilde{B}_t$ by pairing samples in $B_t^{r}$ with DDMs from $\Delta_t$. For a batch $B$, we define its empirical characteristic function in feature space as
\begin{equation}
\hat{\Phi}_{B}(\omega)
=
\frac{1}{|B|}
\sum_{x\in B}
\exp\!\left(j\left\langle \omega, f(x)\right\rangle\right).
\label{eq:empirical_cf}
\end{equation}
We then compare the empirical CFs of the fake batch $B_t^{f}$ and the synthetic fake
batch $\tilde{B}_t$ along sampled frequencies in the complex plane. To focus the matching on the most informative frequencies, we introduce a learnable frequency sampler $q_{\psi}(\omega)$~\cite{ncfm}, implemented as a lightweight MLP that emphasizes frequencies where the two empirical CFs differ the most. The DDC objective is thus written as
\begin{equation}
\begin{array}{rcl}
\mathcal{L}_{\mathrm{DDC}}^{t}
&=&
\displaystyle
\mathbb{E}_{\omega\sim q_{\psi}}
\left[
\left|
\hat{\Phi}_{B_t^{f}}(\omega)
-
\hat{\Phi}_{\tilde{B}_t}(\omega)
\right|
\right]\\[6pt]
&=&
\displaystyle
\int
\left|
\hat{\Phi}_{B_t^{f}}(\omega)
-
\hat{\Phi}_{\tilde{B}_t}(\omega)
\right|
\,dq_{\psi}(\omega).
\end{array}
\label{eq:ddc_loss}
\end{equation}
We optimize DDC with a min--max scheme \cite{ncfm},
\begin{equation}
\min_{\Delta_t}\max_{\psi}\;
\mathcal{L}_{\mathrm{DDC}}^{t},
\label{eq:ddc_minmax}
\end{equation}
where $q_{\psi}$ seeks discrepancy-revealing frequencies, while $\Delta_t$ is updated
to reduce the resulting CF discrepancy. In this way, the learned DDMs progressively
absorb the compact real-to-fake distribution discrepancy of task $t$. After completing task $t$, we freeze the learned DDM bank and append it to memory:
\begin{equation}
\mathcal{M}_{1:t}
=
\mathcal{M}_{1:t-1}\cup\{\Delta_t\}.
\label{eq:memory_update}
\end{equation}
Only the tiny DDM bank is retained, while all raw images from $\mathcal{D}_t$ are
discarded. This yields a compact discrepancy memory without directly storing raw face images.

\subsection{Manifold-Consistent Replay}
A compact discrepancy memory alone is insufficient for continual learning, since replay samples must be jointly trained with current data. Thus, samples synthesized from DDMs should remain statistically aligned with previous distributions while staying on the same energy manifold as current real faces. In practice, we formulate this requirement with an implementable variance-preserving constraint rather than imposing a strict manifold correspondence directly.

\noindent{\textbf{\textit{Variance-Preserving Composition and DDPM-style Formulation.}}} To make the manifold-consistency requirement tractable, we cast it as a
variance-preserving constraint on the composition between current real faces and stored
DDMs. Before composition, each DDM is standardized as
\begin{equation}
\hat{d}_{s,k}
=
\frac{d_{s,k}-\mu(d_{s,k})}{\sigma(d_{s,k})+\varepsilon},
\label{eq:ddm_standardization}
\end{equation}
where $\mu(\cdot)$ and $\sigma(\cdot)$ denote the mean and standard deviation,
respectively, and $\varepsilon$ is a small constant. This removes task-specific scale
variation accumulated during DDC and makes the discrepancy maps suitable for
variance-controlled replay.

Let $x_t^{r}$ be a real face from the current task and $\hat{d}_{s,k}$ be a
standardized DDM from a previous task $s<t$. Since the inputs are normalized and
Eq.~\eqref{eq:ddm_standardization} enforces zero mean and unit variance on the DDMs, we
approximate
\begin{equation}
\mathbb{E}[x_t^{r}] \approx 0,\qquad
\mathbb{E}[\hat{d}_{s,k}] \approx 0,\qquad
\mathrm{Var}(x_t^{r}) \approx \mathrm{Var}(\hat{d}_{s,k}) \approx \mathbf{I},
\end{equation}
and treat $x_t^{r}$ and $\hat{d}_{s,k}$ as approximately independent. Consider the
generic composition
\begin{equation}
\tilde{x}_{s\rightarrow t}
=
a x_t^{r}+b \hat{d}_{s,k}.
\label{eq:generic_replay}
\end{equation}

To approximately keep the synthesized sample on the same energy manifold as current real faces, we adopt the following implementable variance-preserving surrogate condition:
\begin{equation}
\mathrm{Var}(\tilde{x}_{s\rightarrow t})
\approx
\mathrm{Var}(x_t^{r}).
\label{eq:vp_constraint}
\end{equation}
Under the above assumptions,
\begin{equation}
\mathrm{Var}(\tilde{x}_{s\rightarrow t})
\approx
a^{2}\mathrm{Var}(x_t^{r})
+
b^{2}\mathrm{Var}(\hat{d}_{s,k})
\approx
(a^{2}+b^{2})\mathbf{I}.
\end{equation}
Thus, Eq.~\eqref{eq:vp_constraint} is satisfied when
\begin{equation}
a^{2}+b^{2}=1.
\label{eq:unit_circle_constraint}
\end{equation}
Parameterizing
\begin{equation}
a=\sqrt{\alpha_{\tau}},
\qquad
b=\sqrt{1-\alpha_{\tau}},
\qquad
\alpha_{\tau}\in(0,1),
\end{equation}
we obtain
\begin{equation}
\tilde{x}_{s\rightarrow t}
=
\sqrt{\alpha_{\tau}}\,x_t^{r}
+
\sqrt{1-\alpha_{\tau}}\,\hat{d}_{s,k},
\label{eq:vp_replay}
\end{equation}
which shares the same variance-preserving algebraic form as DDPM-style forward mixing~\cite{ddpm}. In our case, this formulation is not introduced as a generative prior; rather, it follows from the variance-preserving requirement imposed by manifold-consistent replay. In practice, we sample $\alpha_{\tau}$ from a truncated cosine schedule~\cite{ddpm2} to preserve facial semantics while injecting discrepancy information.

\noindent{\textbf{\textit{Replay Synthesis with Current Real Faces.}}} At task $t$, MCR generates replay fake samples for each previous task $s<t$ by composing current real faces with the stored DDM bank $\Delta_s$ under Eq.~\eqref{eq:vp_replay}:
\begin{equation}
\tilde{x}_{s\rightarrow t}
=
\sqrt{\alpha_{\tau}}\,x_t^{r}
+
\sqrt{1-\alpha_{\tau}}\,\hat{d}_{s,k},
\qquad
x_t^{r}\sim\mathcal{D}_t^{r},\;
d_{s,k}\sim\Delta_s.
\label{eq:replay_sampling}
\end{equation}
All synthesized samples are assigned the fake label and collected into the replay set
\begin{equation}
\tilde{\mathcal{D}}_{1:t-1\rightarrow t}^{f}
=
\bigcup_{s=1}^{t-1}
\left\{
\tilde{x}_{s\rightarrow t}
\right\}.
\label{eq:replay_set}
\end{equation}
These replay fake samples are then jointly trained with the current-task real and fake samples. Since each stored DDM can be paired with many different real faces, MCR yields substantially more diverse replay than fixed-sample storage while reducing identity leakage risk relative to directly storing historical face images. This design assumes that current-stage real faces remain semantically compatible carriers for historical forgery cues, and a carrier-dependence analysis is provided in the supplementary material.
\begin{table*}[t]
\centering
\setlength{\tabcolsep}{3pt}
\renewcommand{\arraystretch}{0.5}
\scriptsize
\caption{Main results on two widely used CFFD benchmarks. Following their original settings for fair comparison, P1 reports ACC while P2 reports AUC. ``Privacy'' indicates whether the method avoids directly storing raw historical face images. $^{\star}$ denotes directly using our replay for mixed training, while $^{\dagger}$ denotes combining our replay with the strong baseline SUR-LID. Best and second-best AA/AF results at each stage are highlighted in bold and underlined, respectively.}
\vspace{-0.3cm}
\resizebox{\textwidth}{!}{%
\begin{tabular}{l c c c | c c c c c c | c c c c c c}
\toprule
\multirow{2}{*}{Method} & \multirow{2}{*}{Privacy} & \multirow{2}{*}{Replays} & \multirow{2}{*}{Task} & \multicolumn{6}{c|}{P1 (ACC$\uparrow$)} & \multicolumn{6}{c}{P2 (AUC$\uparrow$)} \\
\cmidrule(lr){5-10} \cmidrule(lr){11-16}
& & & & FF++ & DFDCP & DFD & CDF2 & AA & AF$\downarrow$ & Hybrid & FR & FS & EFS & AA & AF$\downarrow$ \\
\midrule

\multirow{4}{*}{\shortstack{LwF\\TPAMI'17~\cite{lwf}}}
& \multirow{4}{*}{Yes}
& \multirow{4}{*}{0}
& T1 & 95.52 & -- & -- & -- & 95.52 & -- & 97.00 & -- & -- & -- & \textbf{97.00} & -- \\
&  &  & T2 & 87.83 & 81.57 & -- & -- & 84.70 & 7.69 & 88.76 & 88.45 & -- & -- & 88.61 & 8.24 \\
&  &  & T3 & 76.16 & 41.78 & 96.36 & -- & 71.43 & 29.58 & 84.07 & 80.99 & 96.44 & -- & 87.24 & 10.20 \\
&  &  & T4 & 67.34 & 67.43 & 84.05 & 87.90 & 76.68 & 18.21 & 78.73 & 56.73 & 93.67 & 92.82 & 79.32 & 17.59 \\
\midrule

\multirow{4}{*}{\shortstack{DER\\CVPR'21~\cite{der}}}
& \multirow{4}{*}{No}
& \multirow{4}{*}{500}
& T1 & 95.45 & -- & -- & -- & 95.45 & -- & 97.00 & -- & -- & -- & \textbf{97.00} & -- \\
&  &  & T2 & 73.86 & 78.56 & -- & -- & 76.21 & 21.59 & 59.03 & 99.73 & -- & -- & 79.38 & 37.97 \\
&  &  & T3 & 60.36 & 55.92 & 95.49 & -- & 70.59 & 28.87 & 68.15 & 19.68 & 97.94 & -- & 61.92 & 54.45 \\
&  &  & T4 & 53.28 & 50.53 & 75.06 & 85.24 & 66.03 & 30.21 & 56.79 & 59.83 & 65.36 & 100.00 & 70.50 & 37.56 \\
\midrule

\multirow{4}{*}{\shortstack{CoReD\\MM'21~\cite{cored}}}
& \multirow{4}{*}{No}
& \multirow{4}{*}{500}
& T1 & 95.50 & -- & -- & -- & 95.50 & -- & 96.65 & -- & -- & -- & 96.65 & -- \\
&  &  & T2 & 92.94 & 87.61 & -- & -- & 90.28 & 2.56 & 93.55 & 79.88 & -- & -- & 86.72 & \underline{3.10} \\
&  &  & T3 & 86.84 & 81.07 & 95.22 & -- & 87.71 & 7.60 & 89.07 & 79.29 & 86.05 & -- & 84.80 & 4.09 \\
&  &  & T4 & 74.08 & 76.59 & 93.41 & 80.78 & 81.22 & 11.42 & 84.54 & 64.29 & 84.17 & 92.63 & 81.41 & 9.86 \\
\midrule

\multirow{4}{*}{\shortstack{DFIL\\MM'23~\cite{dfil}}}
& \multirow{4}{*}{No}
& \multirow{4}{*}{500}
& T1 & 95.67 & -- & -- & -- & 95.67 & -- & 96.46 & -- & -- & -- & 96.46 & -- \\
&  &  & T2 & 93.15 & 88.87 & -- & -- & 91.01 & 2.52 & 55.74 & 99.75 & -- & -- & 77.75 & 40.72 \\
&  &  & T3 & 90.30 & 85.42 & 94.67 & -- & 90.13 & 4.41 & 60.71 & 66.49 & 99.03 & -- & 75.41 & 34.51 \\
&  &  & T4 & 86.28 & 79.53 & 92.36 & 83.81 & 85.49 & 7.01 & 50.83 & 95.56 & 70.81 & 99.96 & 79.29 & 26.01 \\
\midrule

\multirow{4}{*}{\shortstack{DMP\\MM'23~\cite{dmp}}}
& \multirow{4}{*}{Yes}
& \multirow{4}{*}{0}
& T1 & 95.96 & -- & -- & -- & 95.96 & -- & 96.70 & -- & -- & -- & 96.70 & -- \\
&  &  & T2 & 92.71 & 89.72 & -- & -- & 91.22 & 3.25 & 78.99 & 95.12 & -- & -- & 87.06 & 17.71 \\
&  &  & T3 & 92.64 & 86.09 & 94.84 & -- & 91.19 & \underline{3.48} & 65.74 & 80.27 & 90.26 & -- & 78.76 & 22.91 \\
&  &  & T4 & 91.61 & 84.86 & 91.81 & 91.67 & 89.99 & \underline{4.08} & 55.41 & 76.50 & 78.43 & 95.51 & 76.46 & 23.91 \\
\midrule

\multirow{4}{*}{\shortstack{HDP\\IJCV'24~\cite{hdp}}}
& \multirow{4}{*}{Yes}
& \multirow{4}{*}{500}
& T1 & 95.13 & -- & -- & -- & 95.13 & -- & 96.71 & -- & -- & -- & 96.71 & -- \\
&  &  & T2 & 85.28 & 86.42 & -- & -- & 85.85 & 9.85 & 67.41 & 95.45 & -- & -- & 81.43 & 29.30 \\
&  &  & T3 & 74.94 & 77.13 & 95.92 & -- & 82.66 & 14.74 & 63.00 & 71.35 & 95.09 & -- & 76.48 & 28.91 \\
&  &  & T4 & 70.18 & 68.14 & 81.37 & 87.29 & 76.75 & 19.26 & 59.89 & 70.06 & 89.34 & 93.73 & 78.26 & 22.65 \\
\midrule

\multirow{4}{*}{\shortstack{SUR-LID\\CVPR'25~\cite{surlid}}}
& \multirow{4}{*}{No}
& \multirow{4}{*}{500}
& T1 & 95.72 & -- & -- & -- & 95.72 & -- & 96.85 & -- & -- & -- & 96.85 & -- \\
&  &  & T2 & 92.42 & 88.97 & -- & -- & 90.70 & 3.30 & 82.91 & 92.42 & -- & -- & 87.66 & 13.94 \\
&  &  & T3 & 90.87 & 82.52 & 96.86 & -- & 90.08 & 5.65 & 90.50 & 96.26 & 97.94 & -- & \underline{94.90} & \underline{1.26} \\
&  &  & T4 & 89.58 & 83.19 & 89.77 & 91.42 & 88.49 & 6.34 & 87.90 & 96.79 & 93.56 & 99.07 & \underline{94.33} & 2.99 \\
\midrule

\multirow{4}{*}{\shortstack{KAN-CFD\\AAAI'26~\cite{kancfd}}}
& \multirow{4}{*}{Yes}
& \multirow{4}{*}{500}
& T1 & 95.57 & -- & -- & -- & 95.57 & -- & 96.80 & -- & -- & -- & 96.80 & -- \\
&  &  & T2 & 92.40 & 90.97 & -- & -- & 91.69 & 3.17 & 93.31 & 91.76 & -- & -- & \underline{92.54} & \underline{3.49} \\
&  &  & T3 & 90.55 & 86.13 & 96.33 & -- & 91.00 & 4.93 & 89.64 & 90.96 & 95.03 & -- & 91.88 & 3.98 \\
&  &  & T4 & 88.15 & 85.62 & 89.25 & 90.48 & 88.38 & 6.62 & 89.64 & 88.16 & 91.29 & 99.86 & 92.24 & 4.83 \\
\midrule
\multirow{4}{*}{\shortstack{\textbf{\textit{DDC-MIR$^{\star}$}}\\\textbf{\textit{(Ours)}}}}
& \multirow{4}{*}{Yes}
& \multirow{4}{*}{50}
& T1 & 96.01 & -- & -- & -- & \textbf{96.01} & -- & 96.90 & -- & -- & -- & \underline{96.90} & -- \\
&  &  & T2 & 93.76 & 91.18 & -- & -- & \underline{92.47} & \underline{2.25} & 92.85 & 92.07 & -- & -- & 92.46 & 4.05 \\
&  &  & T3 & 92.45 & 86.58 & 97.44 & -- & \underline{92.16} & 4.08 & 87.02 & 94.95 & 96.80 & -- & 92.92 & 3.50 \\
&  &  & T4 & 91.80 & 86.03 & 93.79 & 92.71 & \underline{91.08} & 4.34 & 90.79 & 93.36 & 93.09 & 98.90 & 94.04 & \underline{2.84} \\
\midrule

\multirow{4}{*}{\shortstack{\textbf{\textit{DDC-MIR$^{\dagger}$}}\\\textbf{\textit{(Ours)}}}}
& \multirow{4}{*}{Yes}
& \multirow{4}{*}{50}
& T1 & 96.00 & -- & -- & -- & \underline{96.00} & -- & 97.00 & -- & -- & -- & \textbf{97.00} & -- \\
&  &  & T2 & 94.93 & 91.17 & -- & -- & \textbf{93.05} & \textbf{1.07} & 93.95 & 93.60 & -- & -- & \textbf{93.78} & \textbf{3.05} \\
&  &  & T3 & 92.97 & 89.26 & 96.99 & -- & \textbf{93.07} & \textbf{2.47} & 92.43 & 96.04 & 97.64 & -- & \textbf{95.37} & \textbf{1.07} \\
&  &  & T4 & 91.80 & 88.64 & 95.18 & 93.02 & \textbf{92.16} & \textbf{2.85} & 90.72 & 96.99 & 94.22 & 98.99 & \textbf{95.23} & \textbf{2.10} \\
\bottomrule
\end{tabular}%
}
\label{tab:main_results}
\vspace{-0.2cm}
\end{table*}

\subsection{Overall Training Objective and Optimization}
After completing task $t$, we freeze the extractor and condense the current-task real-to-fake discrepancy into a DDM bank by solving
\begin{equation}
\min_{\Delta_t}\max_{\psi}\;
\mathcal{L}_{\mathrm{DDC}}^{t},
\label{eq:overall_objective}
\end{equation}
where $\psi$ parameterizes the learnable frequency sampler and $\Delta_t$ is the DDM bank of task $t$. At the next task $t{+}1$, the detector $g_{\theta}$ is trained on both current-task data and replay data synthesized from memory $\mathcal{M}_{1:t}$. Let
$\ell_{\mathrm{cls}}$ denote the binary cross-entropy loss, with label $y=0$ for real
faces and $y=1$ for fake faces. The current-task detection loss is
\begin{equation}
\mathcal{L}_{\mathrm{cur}}^{t+1}
=
\mathbb{E}_{(x,y)\sim\mathcal{D}_{t+1}}
\left[
\ell_{\mathrm{cls}}(g_{\theta}(x),y)
\right],
\label{eq:l_cur}
\end{equation}
and the replay loss is
\begin{equation}
\mathcal{L}_{\mathrm{rep}}^{t+1}
=
\mathbb{E}_{\tilde{x}\sim\tilde{\mathcal{D}}_{1:t\rightarrow t+1}^{f}}
\left[
\ell_{\mathrm{cls}}(g_{\theta}(\tilde{x}),1)
\right].
\label{eq:l_rep}
\end{equation}
The detector at task $t{+}1$ is then optimized by
\begin{equation}
\min_{\theta}\;
\mathcal{L}_{\mathrm{cur}}^{t+1}
+
\lambda_{\mathrm{rep}}\mathcal{L}_{\mathrm{rep}}^{t+1},
\label{eq:l_train}
\end{equation}
where $\lambda_{\mathrm{rep}}$ balances replay training. In practice, after task $t$ is learned, we freeze its feature extractor and optimize $\Delta_t$ and $\psi$ with $\mathcal{L}_{\mathrm{DDC}}^{t}$. The learned DDM bank $\Delta_t$ is appended to memory and unfolded on current real faces at task $t{+}1$ to synthesize replay samples for mixed training. In this way, the memory cost grows only with a tiny set of DDMs rather than with historical face images.

\section{Experiments}
\begin{table*}[t]
\centering
\setlength{\tabcolsep}{3.6pt}
\renewcommand{\arraystretch}{1.08}
\small
\caption{Generality of the proposed replay mechanism across replay-based continual forgery detectors. Gray rows replace the original replay module of the method above with our replay. P1 reports ACC and P2 reports AUC following the original benchmarks. Bold numbers indicate that replacing the original replay with ours improves the corresponding result.}
\vspace{-0.3cm}
\resizebox{\textwidth}{!}{%
\begin{tabular}{l c c | c c c c c c c | c c c c c c c}
\toprule
\multirow{3}{*}{Method} & \multirow{3}{*}{Privacy} & \multirow{3}{*}{Replays}
& \multicolumn{7}{c|}{P1 (ACC$\uparrow$)} & \multicolumn{7}{c}{P2 (AUC$\uparrow$)} \\
\cmidrule(lr){4-10} \cmidrule(lr){11-17}
& & & FF++ & \multicolumn{2}{c}{DFDCP} & \multicolumn{2}{c}{DFD} & \multicolumn{2}{c|}{CDF2}
& Hybrid & \multicolumn{2}{c}{FR} & \multicolumn{2}{c}{FS} & \multicolumn{2}{c}{EFS} \\
\cmidrule(lr){5-6} \cmidrule(lr){7-8} \cmidrule(lr){9-10}
\cmidrule(lr){12-13} \cmidrule(lr){14-15} \cmidrule(lr){16-17}
& & & AA & AA & AF$\downarrow$ & AA & AF$\downarrow$ & AA & AF$\downarrow$
& AA & AA & AF$\downarrow$ & AA & AF$\downarrow$ & AA & AF$\downarrow$ \\
\midrule

DER~\cite{der} & No & 500
& 95.45 & 76.21 & 21.59 & 70.59 & 28.87 & 66.03 & 30.21
& 97.00 & 79.38 & 37.97 & 61.92 & 54.45 & 70.50 & 37.56 \\
\rowcolor{gray!15}
+ Ours & \textbf{Yes} & \textbf{50}
& 95.42 & \textbf{86.95} & \textbf{8.93} & \textbf{86.69} & \textbf{8.69} & \textbf{81.87} & \textbf{12.62}
& 96.37 & \textbf{86.86} & \textbf{18.53} & \textbf{83.01} & \textbf{19.24} & \textbf{81.08} & \textbf{18.93} \\
\midrule

CoReD~\cite{cored} & No & 500
& 95.50 & 90.28 & 2.56 & 87.71 & 7.60 & 81.22 & 11.42
& 96.65 & 86.72 & 3.10 & 84.80 & 4.09 & 81.41 & 9.86 \\
\rowcolor{gray!15}
+ Ours & \textbf{Yes} & \textbf{50}
& 95.07 & 90.04 & \textbf{2.32} & \textbf{89.85} & \textbf{4.42} & \textbf{85.46} & \textbf{6.16}
& 96.43 & \textbf{86.78} & \textbf{2.48} & 84.53 & 4.24 & \textbf{83.24} & \textbf{7.31} \\
\midrule

DFIL~\cite{dfil} & No & 500
& 95.67 & 91.01 & 2.52 & 90.13 & 4.41 & 85.49 & 7.01
& 96.46 & 77.75 & 40.72 & 75.41 & 34.51 & 79.29 & 26.01 \\
\rowcolor{gray!15}
+ Ours & \textbf{Yes} & \textbf{50}
& 95.16 & \textbf{91.24} & \textbf{2.50} & \textbf{90.70} & \textbf{3.61} & \textbf{87.12} & \textbf{5.16}
& 96.28 & \textbf{89.79} & \textbf{7.16} & \textbf{86.00} & \textbf{11.00} & \textbf{84.58} & \textbf{12.32} \\
\midrule

HDP~\cite{hdp} & Yes & 500
& 95.13 & 85.85 & 9.85 & 82.66 & 14.74 & 76.75 & 19.26
& 96.71 & 81.43 & 29.30 & 76.48 & 28.91 & 78.26 & 22.65 \\
\rowcolor{gray!15}
+ Ours & \textbf{Yes} & \textbf{50}
& \textbf{95.31} & \textbf{87.35} & \textbf{6.92} & \textbf{87.53} & \textbf{7.37} & \textbf{83.40} & \textbf{10.58}
& 96.23 & \textbf{90.87} & \textbf{10.14} & \textbf{89.39} & \textbf{9.82} & \textbf{87.54} & \textbf{10.45} \\
\midrule

SUR-LID~\cite{surlid} & No & 500
& 95.72 & 90.70 & 3.30 & 90.08 & 5.65 & 88.49 & 6.34
& 96.85 & 87.66 & 13.94 & 94.90 & 1.26 & 94.33 & 2.99 \\
\rowcolor{gray!15}
+ Ours & \textbf{Yes} & \textbf{50}
& \textbf{96.00} & \textbf{93.05} & \textbf{1.07} & \textbf{93.07} & \textbf{2.47} & \textbf{92.16} & \textbf{2.85}
& \textbf{97.00} & \textbf{93.78} & \textbf{3.05} & \textbf{95.37} & \textbf{1.07} & \textbf{95.23} & \textbf{2.10} \\
\bottomrule
\end{tabular}%
}
\label{tab:plugplay}
\end{table*}
\subsection{Datasets and Implementation Details}
\noindent{\textbf{\textit{Datasets.}}} We conduct experiments on a diverse set of face forgery benchmarks. Specifically, we use FaceForensics++ (FF++)~\cite{ffpp}, Deepfake Detection (DFD)~\cite{dfd}, Deepfake Detection Challenge Preview (DFDC-P)~\cite{dfdc}, Celeb-DF v2 (CDF2)~\cite{celeb}, and DF40~\cite{df40}. Among them, FF++ contains manipulations from DF, F2F~\cite{f2f}, FS, and NT~\cite{NT}, and is thus treated as a hybrid-forgery dataset, while DF40~\cite{df40} provides a broad pool of forgery categories. 

\noindent{\textbf{\textit{Incremental Protocols.}}} We evaluate under two continual settings. Protocol 1 (P1): Dataset Incremental follows prior replay-based CFFD benchmarks and uses the task sequence \{FF++, DFDC-P, DFD, CDF2\}, where both real and fake data vary across stages. Protocol 2 (P2): Forgery-Type Incremental focuses on newly emerging forgery categories with consistent real-face carriers, using the sequence \{Hybrid (FF++), FR (MCNet~\cite{mcnet}), FS (BlendFace~\cite{blendface}), EFS (StyleGAN3~\cite{stylegan3})\}, where the latter three are selected from DF40~\cite{df40}.

\noindent{\textbf{\textit{Metrics.}}} Since P1 and P2 follow the public DFIL and SUR-LID benchmarks, respectively, we report the same frame-level ACC and AUC as in their original settings for fair comparison. We also report average forgetting (AF), defined as the average drop in each task's performance from when it is learned to its final performance.

\noindent{\textbf{\textit{Implementation Details.}}} The proposed method is built on DeepFakeBench~\cite{deepfakebench} and resizes all inputs to $256\times256$. EfficientNet-B4~\cite{effnet} is used as the detector backbone. Training is performed with Adam~\cite{adam} using $\beta_1=0.9$, $\beta_2=0.999$, weight decay $5\times10^{-4}$, and an initial learning rate of $2\times10^{-4}$ with step decay. Each task is trained for 20 epochs with a global batch size of 256. Following the benchmark setting, we sample 8 frames per video for training. For continual replay, we store 50 DDMs for each task. During the DDC, only the fake class is distilled into 50 discrepancy maps per task for 3000 iterations with a learning rate of 0.01. All experiments are conducted on two NVIDIA A6000 GPUs. We follow the standard single-run benchmark protocol; matched-budget fairness, multi-seed robustness, and carrier-dependence analyses are deferred to the supplementary material.
\vspace{-2pt}
\subsection{Comparison Experiments}
We compare with representative continual learning and CFFD baselines, including LwF~\cite{lwf}, DER~\cite{der}, CoReD~\cite{cored}, DFIL~\cite{dfil}, DMP~\cite{dmp}, HDP~\cite{hdp}, SUR-LID~\cite{surlid}, and KAN-CFD~\cite{kancfd}.
Following Table~\ref{tab:main_results}, LwF, HDP, KAN-CFD, and our method are marked as \emph{Yes} in the ``Privacy'' column because they avoid directly storing raw face images. We report our method in two settings. \textbf{DDC-MIR$^{\star}$} directly uses the proposed replay for mixed training, evaluating the replay itself without additional continual operations. \textbf{DDC-MIR$^{\dagger}$} further integrates our replay module into the strong baseline, SUR-LID, showing the best performance attainable.

Table~\ref{tab:main_results} shows that our method achieves the best performance under both protocols. On the dataset-incremental benchmark P1, \textbf{DDC-MIR$^{\star}$} already delivers the best standalone final result, reaching 91.08 AA with 4.34 AF, outperforming all prior baselines. Since this setting simply mixes current data with our replay, the result highlights the strength of our replay itself. When further combined with SUR-LID, \textbf{DDC-MIR$^{\dagger}$} pushes the final result to 92.16 AA with 2.85 AF, establishing a clear new state of the art on P1.

A similar trend is observed on the forgery-type incremental benchmark P2. \textbf{DDC-MIR$^{\star}$} already reduces forgetting to 2.84, surpassing all existing baselines in AF. Moreover, \textbf{DDC-MIR$^{\dagger}$} achieves the best overall result on P2, reaching 95.23 AA with 2.10 AF, outperforming the previous strongest method SUR-LID. Notably, this integrated version yields state-of-the-art performance at every continual stage, showing that our method is the strongest replay replacement for advanced continual forgery detection frameworks. Supplementary material further reports matched-budget fairness and multi-seed robustness analyses, which support the same trend.

\subsection{Replay Mechanism Generality}
To verify that our method is a generally stronger replay mechanism rather than a framework-specific design, we replace the original replay module in representative replay-based CFFD methods, including DER~\cite{der}, CoReD~\cite{cored}, DFIL~\cite{dfil}, HDP~\cite{hdp}, and SUR-LID~\cite{surlid}. 

As shown in Table~\ref{tab:plugplay}, our replay consistently improves all compared methods under both continual protocols. On P1, replacing the original replay with ours yields clear gains in final AA and substantially reduces AF for every method. The improvements are particularly large for methods with weaker replay, such as DER and HDP, while even strong baselines such as DFIL and SUR-LID continue to benefit. A similar trend is observed on P2, where replacing the original replay again improves the final results across all compared frameworks. This shows that our method is not only a strong standalone continual learning solution, but also a generally applicable and plug-and-play replay module with strong mechanism-level generality across different continual forgery detection frameworks.

\subsection{Historical-Information Preserving Analysis}
\begin{figure*}[t]
    \centering
    \includegraphics[width=\textwidth]{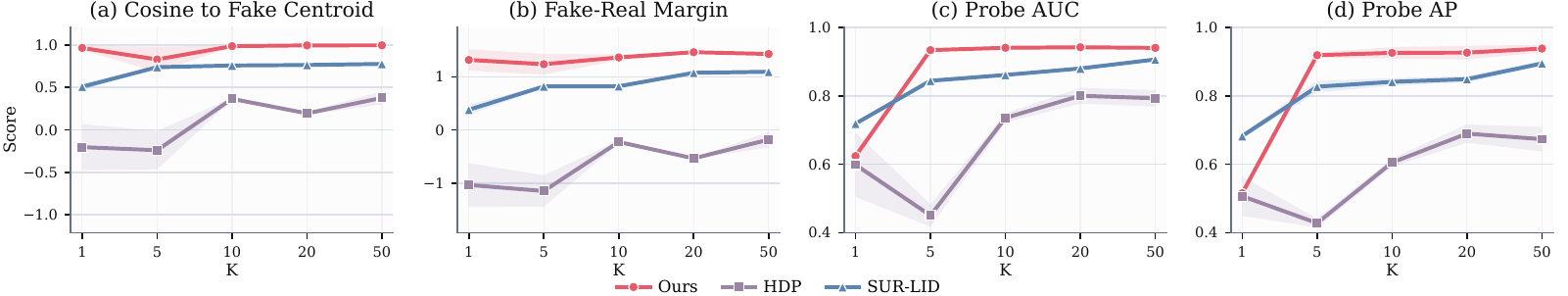}
    \caption{Analysis of historical information encoded in replay samples under different replay budgets $K\in\{1,5,10,20,50\}$. (a) Cosine to Fake Centroid, (b) Fake-Real Margin, (c) Probe AUC, and (d) Probe AP.}
    \label{fig:exp3}
    % \vspace{-0.2cm}
    
\end{figure*}
To better understand what different replay mechanisms actually preserve, we analyze replay samples in the frozen feature space of the task-0 detector. We compare our method with HDP~\cite{hdp} and SUR-LID~\cite{surlid} under different replay budgets $K$.

\noindent{\textbf{\textit{Feature-Space Distribution Analysis.}}}
We first examine whether replay samples are globally aligned with the historical fake distribution in the old-task feature space using \emph{Cosine to Fake Centroid} and \emph{Fake-Real Margin}. For SUR-LID, the replay set consists of selected replay images, whereas for HDP and our method it consists of synthesized fake samples. Fig.~\ref{fig:exp3}(a)--(b) shows that our method yields the strongest set-level alignment with the historical fake distribution. Its cosine-to-fake score approaches 1.0 and remains stable as $K$ increases, while its fake-real margin stays clearly positive and consistently the largest. SUR-LID is better than HDP, but remains consistently below our method. This indicates that our replay more faithfully captures the geometry of the previous fake distribution.

\noindent{\textbf{\textit{Linear Probe Re-training Analysis.}}}
We next test whether replay samples can support the recovery of old-task classification ability. We train a linear probe on the frozen extractor using only held-out old real samples and replay fake samples, and evaluate it on the FF++ test set. Fig.~\ref{fig:exp3}(c)--(d) shows that our method achieves the best overall Probe AUC and Probe AP, overtaking SUR-LID once a modest number of maps are available and remaining the strongest for $K\geq5$. HDP is consistently much weaker across the entire budget range. Together, these results show that the replay synthesized from DDMs carries sufficient discriminative structure to recover the old-task classifier.

\subsection{Replay-Level Privacy Analysis}
As a replay-level privacy analysis, we stop immediately after task-0 training on FF++ and compare the materialized replay of different replay methods. This isolates whether replay itself exposes historical facial information. We consider two complementary views: \emph{identity-level linkability}, measured by ArcFace-based~\cite{arcface} linkage and membership inference metrics, and \emph{image-level visual similarity}, measured by LPIPS-based~\cite{lpips} nearest-neighbor and membership inference metrics. These measurements are privacy proxies.

\noindent\textbf{\textit{Identity-level linkability.}}
We first ask whether replay can be linked back to historical identities. To this end, we use a frozen ArcFace~\cite{arcface} encoder and report ArcFace R1-Link at FAR=1\% and 0.1\%, together with Replay-MIA-ArcFace AUC. Lower values indicate weaker identity linkability. As shown in Table~\ref{tab:privacy}, our method performs best on all three metrics. In particular, ArcFace R1-Link@0.1\% FAR drops to 3.91\%, compared with 5.86\% for HDP and over 24\% for DFIL and SUR-LID. Replay-MIA-ArcFace AUC is also lowest for our method (43.36\%), showing that our replay is the hardest to associate with historical identities.

\noindent\textbf{\textit{Image-level visual similarity.}}
We further analyze whether replay remains visually traceable to historical real faces. We report the nearest-neighbor LPIPS~\cite{lpips} distance from replay to the historical real-face gallery and Replay-MIA-LPIPS AUC. Higher LPIPS and lower LPIPS-AUC indicate weaker visual reversibility. Our method again performs best, achieving the largest LPIPS value (0.88) and the lowest Replay-MIA-LPIPS AUC (37.60\%). Compared with HDP, DFIL, and SUR-LID, our replay is visually the farthest from historical faces and is the least vulnerable to perceptual membership inference. Overall, these results suggest that our replay is less identity-linkable and less visually reversible than existing methods under the evaluated proxies.

\begin{table}[t]
\centering
\setlength{\tabcolsep}{3pt}
\renewcommand{\arraystretch}{1.08}
\small
\caption{Replay-level privacy analysis. R1@1/R1@0.1: ArcFace R1-Link at FAR=1\%/0.1\%; MIA-AUC: Replay-MIA-ArcFace AUC; LPIPS: NN-LPIPS mean; LPIPS AUC: Replay-MIA-LPIPS AUC. Lower is better except for LPIPS.}
\vspace{-0.4cm}
\resizebox{\columnwidth}{!}{%
\begin{tabular}{lccc|cc}
\toprule
\multirow{2}{*}{Method} & \multicolumn{3}{c|}{Identity~\cite{arcface}} & \multicolumn{2}{c}{Visual~\cite{lpips}} \\
\cmidrule(lr){2-4} \cmidrule(lr){5-6}
& R1@1$\downarrow$ & R1@0.1$\downarrow$ & MIA-AUC$\downarrow$ & LPIPS$\uparrow$ & LPIPS AUC$\downarrow$ \\
\midrule
HDP~\cite{hdp} & \underline{16.80} & \underline{5.86} & \underline{46.09} & \underline{0.50} & 49.56 \\
DFIL~\cite{dfil} & 47.27 & 27.34 & 67.38 & 0.47 & 45.34 \\
SUR-LID~\cite{surlid} & 44.14 & 24.22 & 65.82 & 0.47 & \underline{41.24} \\
Ours & \textbf{16.41} & \textbf{3.91} & \textbf{43.36} & \textbf{0.88} & \textbf{37.60} \\
\bottomrule
\end{tabular}
}
\label{tab:privacy}
\vspace{-0.25cm}
\end{table}
\begin{table}[t]
\centering
\setlength{\tabcolsep}{3pt}
\renewcommand{\arraystretch}{1.08}
\small
\caption{Ablation study on Protocol 1. We report the average accuracy (AA) and average forgetting (AF) after each task to analyze the contribution of DDC, DDM standardization, and MCR. Best results are in bold, and the second are underlined.}
\vspace{-0.35cm}
\begin{tabular}{l|cc|cc|cc|cc}
\toprule
\multirow{2}{*}{Variant} & \multicolumn{2}{c|}{FF++} & \multicolumn{2}{c|}{DFDCP} & \multicolumn{2}{c|}{DFD} & \multicolumn{2}{c}{CDF2} \\
\cmidrule(lr){2-3} \cmidrule(lr){4-5} \cmidrule(lr){6-7} \cmidrule(lr){8-9}
& AA$\uparrow$ & AF$\downarrow$ & AA$\uparrow$ & AF$\downarrow$ & AA$\uparrow$ & AF$\downarrow$ & AA$\uparrow$ & AF$\downarrow$ \\
\midrule
Full & \underline{96.01} & - & \textbf{92.47} & \textbf{2.25} & \textbf{92.16} & \textbf{4.08} & \textbf{91.08} & \textbf{4.34} \\
w/o DDC & \textbf{96.05} & - & 86.32 & 14.50 & 83.24 & 16.31 & 79.82 & 18.36 \\
w/o DDM Std. & \textbf{96.05} & - & \underline{92.35} & \underline{3.25} & \underline{90.08} & \underline{6.54} & \underline{86.50} & \underline{9.86} \\
w/o MCR & 96.00 & - & 88.70 & 9.75 & 87.38 & 10.12 & 83.47 & 13.68 \\
\bottomrule
\end{tabular}
\label{tab:ablation}
\vspace{-0.2cm}
\end{table}

\subsection{Replay and Distribution Visualization}

We compare two representative replay paradigms with our method after training on FF++: SUR-LID~\cite{surlid} as a selection-based replay method and HDP~\cite{hdp} as a perturbation-based replay method.

\begin{figure}[t]
    \centering
    \includegraphics[width=\linewidth]{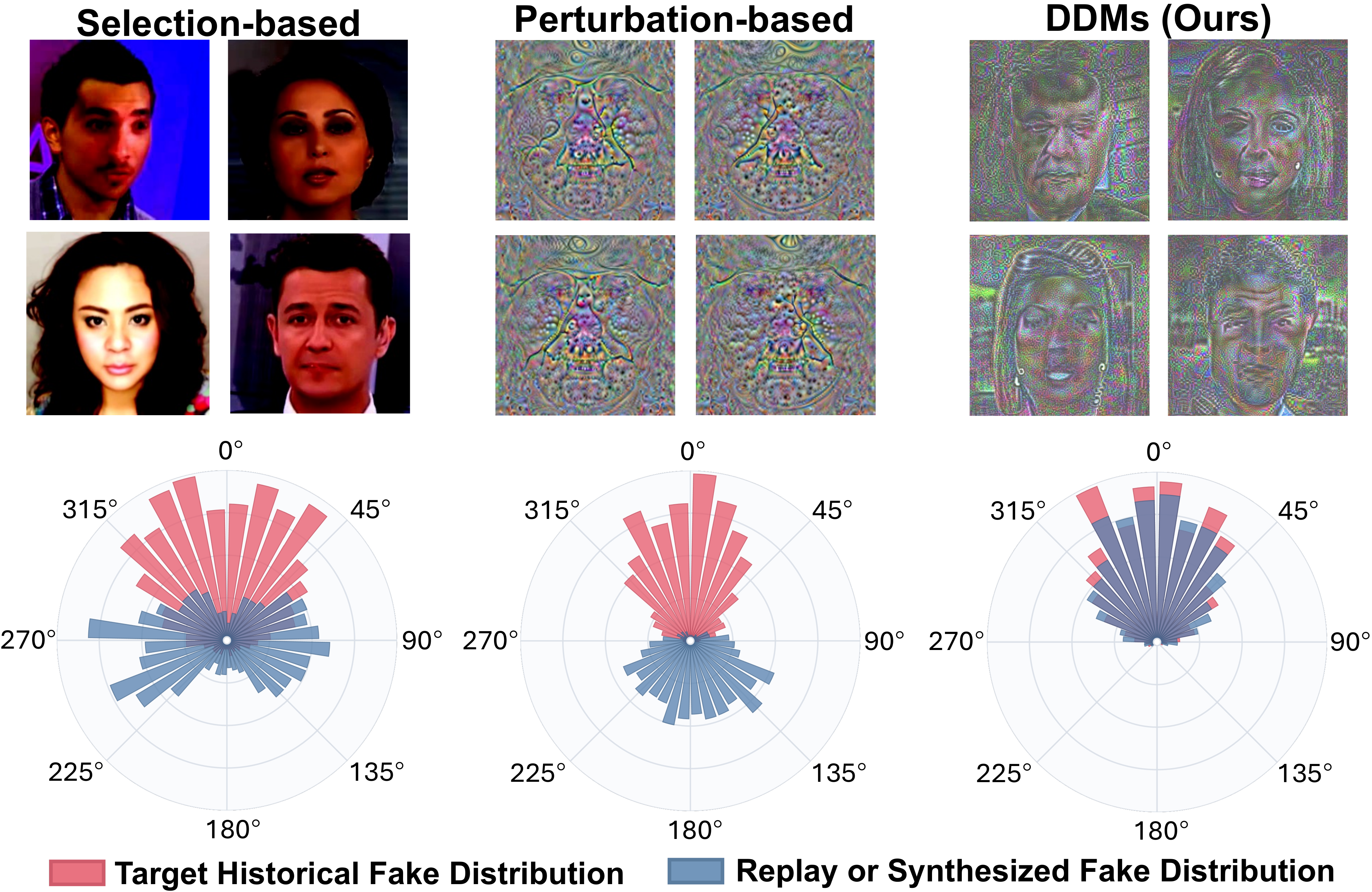}
    \caption{Visualization of replay samples and distribution consistency after task-0 training on FF++. Top: representative replay samples from selection-based replay, perturbation-based replay, and our DDM replay. Bottom: visualization of characteristic functions, where angle denotes phase and radial magnitude denotes amplitude.}
    \label{fig:dis_vis}
    \vspace{-0.3cm}
\end{figure}

\noindent\textbf{\textit{Replay sample visualization.}}
As shown in the top row of Fig.~\ref{fig:dis_vis}, selection-based replay directly retains historical faces, which exposes clear privacy risks. Perturbation-based replay mainly preserves boundary-oriented signals and coarse facial outlines, but lacks rich forgery content. In contrast, our DDM-based replay preserves substantially richer facial information while remaining clearly different from directly stored faces.

\noindent\textbf{\textit{Distribution consistency visualization.}}
The bottom row of Fig.~\ref{fig:dis_vis} compares the target historical fake distribution and the replay distribution through characteristic functions (CFs) visualization. Selection-based replay relies on sparse retained samples and only coarsely approximates the historical fake distribution. Perturbation-based replay is even farther from the target, since it mainly preserves boundary directions rather than the original fake distribution. By contrast, our replay closely follows the target in both phase and amplitude, showing much stronger distribution consistency.
\subsection{Ablation Study}
We perform ablation studies on Protocol 1 to assess the contribution of each component in our framework. \textbf{Full} denotes the complete model. \textbf{w/o DDC} removes the distribution-matching constraint and only supervises the synthesized samples with the fake label. \textbf{w/o DDM Std.} removes the standardization operation in DDM. \textbf{w/o MCR} replaces the proposed MCR with a naive direct-addition strategy for sample synthesis. All variants share the same backbone, optimization setting, and evaluation protocol. We report the average accuracy (AA) and average forgetting (AF) after each task.

As shown in Table~\ref{tab:ablation}, all variants achieve comparable performance on the first task, indicating that the observed differences mainly stem from continual knowledge preservation rather than single-task fitting. Removing DDC causes the largest degradation, reducing the final AA from 91.08 to 79.82 and increasing the final AF from 4.34 to 18.36, which verifies that effective replay requires explicit historical distribution alignment rather than only fake-label supervision. Replacing MCR with naive direct addition also significantly hurts performance, leading to 83.47 AA and 13.68 AF at the final task, demonstrating that simple feature composition is insufficient for constructing high-quality replay samples. Removing DDM standardization yields the strongest ablated variant, but it still falls behind the full model, especially in long-term forgetting. Overall, these results confirm that DDC is the most critical component, while MCR and DDM standardization further provide complementary gains for stable replay and continual retention.

\section{Conclusion}
We presented a direct discrepancy replay framework for continual face forgery detection. Instead of retaining raw faces or boundary-tied perturbations, our method condenses a surrogate real-to-fake discrepancy into compact distribution discrepancy maps and replays the distribution of prior tasks via variance-preserving composition with current real faces. Extensive experiments show that the proposed replay achieves strong continual detection performance both as a standalone solution and as a replacement for existing replay modules. Replay-level privacy analysis further suggests a reduced identity leakage risk under the evaluated proxies.

\bibliographystyle{ACM-Reference-Format}
\bibliography{sample-base}

@String{Computer = "{IEEE} Computer" }

@String{Springer = "Springer-Verlag" }

@article{hdp,
  title={Continual face forgery detection via historical distribution preserving},
  author={Sun, Ke and Chen, Shen and Yao, Taiping and Sun, Xiaoshuai and Ding, Shouhong and Ji, Rongrong},
  journal={International Journal of Computer Vision},
  volume={133},
  number={3},
  pages={1067--1084},
  year={2025},
  publisher={Springer}
}

@inproceedings{dmp,
  title={Dynamic Mixed-Prototype Model for Incremental Deepfake Detection},
  author={Tian, Jiahe and Yu, Cai and Wang, Xi and Chen, Peng and Xiao, Zihao and Han, Jizhong and Chai, Yesheng},
  booktitle={Proceedings of the 32nd ACM International Conference on Multimedia},
  pages={8129--8138},
  year={2024}
}

@inproceedings{surlid,
  title={Stacking brick by brick: Aligned feature isolation for incremental face forgery detection},
  author={Cheng, Jikang and Yan, Zhiyuan and Zhang, Ying and Hao, Li and Ai, Jiaxin and Zou, Qin and Li, Chen and Wang, Zhongyuan},
  booktitle={Proceedings of the Computer Vision and Pattern Recognition Conference},
  pages={13927--13936},
  year={2025}
}

@inproceedings{cored,
  title={Cored: Generalizing fake media detection with continual representation using distillation},
  author={Kim, Minha and Tariq, Shahroz and Woo, Simon S},
  booktitle={Proceedings of the 29th ACM International Conference on Multimedia},
  pages={337--346},
  year={2021}
}

@inproceedings{gan_gen2,
  title={Stylemask: Disentangling the style space of stylegan2 for neural face reenactment},
  author={Bounareli, Stella and Tzelepis, Christos and Argyriou, Vasileios and Patras, Ioannis and Tzimiropoulos, Georgios},
  booktitle={2023 IEEE 17th international conference on automatic face and gesture recognition (FG)},
  pages={1--8},
  year={2023},
  organization={IEEE}
}

@inproceedings{gan_gen3,
  title={Styleswap: Style-based generator empowers robust face swapping},
  author={Xu, Zhiliang and Zhou, Hang and Hong, Zhibin and Liu, Ziwei and Liu, Jiaming and Guo, Zhizhi and Han, Junyu and Liu, Jingtuo and Ding, Errui and Wang, Jingdong},
  booktitle={European Conference on Computer Vision},
  pages={661--677},
  year={2022},
  organization={Springer}
}

@article{diff2,
  title={Hallo: Hierarchical Audio-Driven Visual Synthesis for Portrait Image Animation},
  author={Xu, Mingwang and Li, Hui and Su, Qingkun and Shang, Hanlin and Zhang, Liwei and Liu, Ce and Wang, Jingdong and Van Gool, Luc and Yao, Yao and Zhu, Siyu},
  journal={arXiv preprint arXiv:2406.08801},
  year={2024}
}

@inproceedings{diff3,
  title={Diffusion video autoencoders: Toward temporally consistent face video editing via disentangled video encoding},
  author={Kim, Gyeongman and Shim, Hajin and Kim, Hyunsu and Choi, Yunjey and Kim, Junho and Yang, Eunho},
  booktitle={Proceedings of the IEEE/CVF Conference on Computer Vision and Pattern Recognition},
  pages={6091--6100},
  year={2023}
}

@inproceedings{sbi,
  title={Detecting deepfakes with self-blended images},
  author={Shiohara, Kaede and Yamasaki, Toshihiko},
  booktitle={Proceedings of the IEEE/CVF Conference on Computer Vision and Pattern Recognition},
  pages={18720--18729},
  year={2022}
}

@article{moeffd,
  title={MoE-FFD: Mixture of Experts for Generalized and Parameter-Efficient Face Forgery Detection},
  author={Kong, Chenqi and Luo, Anwei and Xia, Song and Yu, Yi and Li, Haoliang and Kot, Alex C},
  journal={arXiv preprint arXiv:2404.08452},
  year={2024}
}

@inproceedings{dfil,
  title={Dfil: Deepfake incremental learning by exploiting domain-invariant forgery clues},
  author={Pan, Kun and Yin, Yifang and Wei, Yao and Lin, Feng and Ba, Zhongjie and Liu, Zhenguang and Wang, Zhibo and Cavallaro, Lorenzo and Ren, Kui},
  booktitle={Proceedings of the 31st ACM International Conference on Multimedia},
  pages={8035--8046},
  year={2023}
}

@inproceedings{ffpp,
  title={Faceforensics++: Learning to detect manipulated facial images},
  author={Rossler, Andreas and Cozzolino, Davide and Verdoliva, Luisa and Riess, Christian and Thies, Justus and Nie{\ss}ner, Matthias},
  booktitle={Proceedings of the IEEE/CVF international conference on computer vision},
  pages={1--11},
  year={2019}
}

@inproceedings{f2f,
  title={Face2face: Real-time face capture and reenactment of rgb videos},
  author={Thies, Justus and Zollhofer, Michael and Stamminger, Marc and Theobalt, Christian and Nie{\ss}ner, Matthias},
  booktitle={Proceedings of the IEEE conference on computer vision and pattern recognition},
  pages={2387--2395},
  year={2016}
}

@article{nt,
  title={Deferred neural rendering: Image synthesis using neural textures},
  author={Thies, Justus and Zollh{\"o}fer, Michael and Nie{\ss}ner, Matthias},
  journal={ACM Transactions on Graphics (TOG)},
  volume={38},
  number={4},
  pages={1--12},
  year={2019},
  publisher={ACM New York, NY, USA}
}

@inproceedings{celeb,
  title={Celeb-df: A large-scale challenging dataset for deepfake forensics},
  author={Li, Yuezun and Yang, Xin and Sun, Pu and Qi, Honggang and Lyu, Siwei},
  booktitle={Proceedings of the IEEE/CVF conference on computer vision and pattern recognition},
  pages={3207--3216},
  year={2020}
}

@article{dfd,
  title={Contributing data to deepfake detection research},
  author={Dufour, Nick and Gully, Andrew},
  journal={Google AI Blog},
  volume={1},
  number={2},
  pages={3},
  year={2019}
}

@article{dfdc,
  title={The deepfake detection challenge (dfdc) dataset},
  author={Dolhansky, Brian and Bitton, Joanna and Pflaum, Ben and Lu, Jikuo and Howes, Russ and Wang, Menglin and Ferrer, Cristian Canton},
  journal={arXiv preprint arXiv:2006.07397},
  year={2020}
}

@article{lwf,
  title={Learning without forgetting},
  author={Li, Zhizhong and Hoiem, Derek},
  journal={IEEE transactions on pattern analysis and machine intelligence},
  volume={40},
  number={12},
  pages={2935--2947},
  year={2017},
  publisher={IEEE}
}

@article{ewc,
  title={Overcoming catastrophic forgetting in neural networks},
  author={Kirkpatrick, James and Pascanu, Razvan and Rabinowitz, Neil and Veness, Joel and Desjardins, Guillaume and Rusu, Andrei A and Milan, Kieran and Quan, John and Ramalho, Tiago and Grabska-Barwinska, Agnieszka and others},
  journal={Proceedings of the national academy of sciences},
  volume={114},
  number={13},
  pages={3521--3526},
  year={2017},
  publisher={National Acad Sciences}
}

@article{dgr,
  title={Continual learning with deep generative replay},
  author={Shin, Hanul and Lee, Jung Kwon and Kim, Jaehong and Kim, Jiwon},
  journal={Advances in neural information processing systems},
  volume={30},
  year={2017}
}

@article{er,
  title={On tiny episodic memories in continual learning},
  author={Chaudhry, Arslan and Rohrbach, Marcus and Elhoseiny, Mohamed and Ajanthan, Thalaiyasingam and Dokania, Puneet K and Torr, Philip HS and Ranzato, Marc'Aurelio},
  journal={arXiv preprint arXiv:1902.10486},
  year={2019}
}

@inproceedings{si,
  title={Continual learning through synaptic intelligence},
  author={Zenke, Friedemann and Poole, Ben and Ganguli, Surya},
  booktitle={International conference on machine learning},
  pages={3987--3995},
  year={2017},
  organization={PMLR}
}

@inproceedings{icarl,
  title={icarl: Incremental classifier and representation learning},
  author={Rebuffi, Sylvestre-Alvise and Kolesnikov, Alexander and Sperl, Georg and Lampert, Christoph H},
  booktitle={Proceedings of the IEEE conference on Computer Vision and Pattern Recognition},
  pages={2001--2010},
  year={2017}
}

@inproceedings{der,
  title={Der: Dynamically expandable representation for class incremental learning},
  author={Yan, Shipeng and Xie, Jiangwei and He, Xuming},
  booktitle={Proceedings of the IEEE/CVF conference on computer vision and pattern recognition},
  pages={3014--3023},
  year={2021}
}

@article{df40,
  title={DF40: Toward Next-Generation Deepfake Detection},
  author={Yan, Zhiyuan and Yao, Taiping and Chen, Shen and Zhao, Yandan and Fu, Xinghe and Zhu, Junwei and Luo, Donghao and Yuan, Li and Wang, Chengjie and Ding, Shouhong and others},
  journal={arXiv preprint arXiv:2406.13495},
  year={2024}
}

@inproceedings{effnet,
  title={Efficientnet: Rethinking model scaling for convolutional neural networks},
  author={Tan, Mingxing and Le, Quoc},
  booktitle={International conference on machine learning},
  pages={6105--6114},
  year={2019},
  organization={PMLR}
}

@inproceedings{deepfakebench,
 author = {Yan, Zhiyuan and Zhang, Yong and Yuan, Xinhang and Lyu, Siwei and Wu, Baoyuan},
 booktitle = {Advances in Neural Information Processing Systems},
 editor = {A. Oh and T. Neumann and A. Globerson and K. Saenko and M. Hardt and S. Levine},
 pages = {4534--4565},
 publisher = {Curran Associates, Inc.},
 title = {DeepfakeBench: A Comprehensive Benchmark of Deepfake Detection},
 url = {https://proceedings.neurips.cc/paper_files/paper/2023/file/0e735e4b4f07de483cbe250130992726-Paper-Datasets_and_Benchmarks.pdf},
 volume = {36},
 year = {2023}
}

@article{adam,
  title={Adam: A method for stochastic optimization},
  author={Kingma, Diederik P and Ba, Jimmy},
  journal={arXiv preprint arXiv:1412.6980},
  year={2014}
}

@article{dd,
  title={Dataset distillation},
  author={Wang, Tongzhou and Zhu, Jun-Yan and Torralba, Antonio and Efros, Alexei A},
  journal={arXiv preprint arXiv:1811.10959},
  year={2018}
}

@article{dc,
  title={Dataset condensation with gradient matching},
  author={Zhao, Bo and Mopuri, Konda Reddy and Bilen, Hakan},
  journal={arXiv preprint arXiv:2006.05929},
  year={2020}
}

@inproceedings{dsa,
  title={Dataset condensation with differentiable siamese augmentation},
  author={Zhao, Bo and Bilen, Hakan},
  booktitle={International Conference on Machine Learning},
  pages={12674--12685},
  year={2021},
  organization={PMLR}
}

@inproceedings{mtt,
  title={Dataset distillation by matching training trajectories},
  author={Cazenavette, George and Wang, Tongzhou and Torralba, Antonio and Efros, Alexei A and Zhu, Jun-Yan},
  booktitle={Proceedings of the IEEE/CVF conference on computer vision and pattern recognition},
  pages={4750--4759},
  year={2022}
}

@article{datm,
  title={Towards lossless dataset distillation via difficulty-aligned trajectory matching},
  author={Guo, Ziyao and Wang, Kai and Cazenavette, George and Li, Hui and Zhang, Kaipeng and You, Yang},
  journal={arXiv preprint arXiv:2310.05773},
  year={2023}
}

@inproceedings{dm,
  title={Dataset condensation with distribution matching},
  author={Zhao, Bo and Bilen, Hakan},
  booktitle={Proceedings of the IEEE/CVF winter conference on applications of computer vision},
  pages={6514--6523},
  year={2023}
}

@inproceedings{cafe,
  title={Cafe: Learning to condense dataset by aligning features},
  author={Wang, Kai and Zhao, Bo and Peng, Xiangyu and Zhu, Zheng and Yang, Shuo and Wang, Shuo and Huang, Guan and Bilen, Hakan and Wang, Xinchao and You, Yang},
  booktitle={Proceedings of the IEEE/CVF Conference on Computer Vision and Pattern Recognition},
  pages={12196--12205},
  year={2022}
}

@inproceedings{idm,
  title={Improved distribution matching for dataset condensation},
  author={Zhao, Ganlong and Li, Guanbin and Qin, Yipeng and Yu, Yizhou},
  booktitle={Proceedings of the IEEE/CVF conference on computer vision and pattern recognition},
  pages={7856--7865},
  year={2023}
}

@inproceedings{m3d,
  title={M3d: Dataset condensation by minimizing maximum mean discrepancy},
  author={Zhang, Hansong and Li, Shikun and Wang, Pengju and Zeng, Dan and Ge, Shiming},
  booktitle={Proceedings of the AAAI Conference on Artificial Intelligence},
  volume={38},
  number={8},
  pages={9314--9322},
  year={2024}
}

@inproceedings{ncfm,
  title={Dataset distillation with neural characteristic function: A minmax perspective},
  author={Wang, Shaobo and Yang, Yicun and Liu, Zhiyuan and Sun, Chenghao and Hu, Xuming and He, Conghui and Zhang, Linfeng},
  booktitle={Proceedings of the Computer Vision and Pattern Recognition Conference},
  pages={25570--25580},
  year={2025}
}

@inproceedings{kancfd,
  title={Unifying Locality of KANs and Feature Drift Compensation Projection for Data-free Replay based Continual Face Forgery Detection},
  author={Zhang, Tianshuo and Peng, Siran and Gao, Li and Zhang, Haoyuan and Zhu, Xiangyu and Lei, Zhen},
  booktitle={Proceedings of the AAAI Conference on Artificial Intelligence},
  volume={40},
  number={15},
  pages={12771--12779},
  year={2026}
}

@book{cf,
  title={Probability and measure},
  author={Billingsley, Patrick},
  year={2017},
  publisher={John Wiley \& Sons}
}

@article{gem,
  title={Gradient episodic memory for continual learning},
  author={Lopez-Paz, David and Ranzato, Marc'Aurelio},
  journal={Advances in neural information processing systems},
  volume={30},
  year={2017}
}

@article{replay,
  title={Catastrophic forgetting, rehearsal and pseudorehearsal},
  author={Robins, Anthony},
  journal={Connection Science},
  volume={7},
  number={2},
  pages={123--146},
  year={1995},
  publisher={Taylor \& Francis}
}

@article{ddpm,
  title={Denoising diffusion probabilistic models},
  author={Ho, Jonathan and Jain, Ajay and Abbeel, Pieter},
  journal={Advances in neural information processing systems},
  volume={33},
  pages={6840--6851},
  year={2020}
}

@inproceedings{ddpm2,
  title={Improved denoising diffusion probabilistic models},
  author={Nichol, Alexander Quinn and Dhariwal, Prafulla},
  booktitle={International conference on machine learning},
  pages={8162--8171},
  year={2021},
  organization={PMLR}
}

@inproceedings{mcnet,
  title={Implicit identity representation conditioned memory compensation network for talking head video generation},
  author={Hong, Fa-Ting and Xu, Dan},
  booktitle={Proceedings of the IEEE/CVF international conference on computer vision},
  pages={23062--23072},
  year={2023}
}

@inproceedings{blendface,
  title={Blendface: Re-designing identity encoders for face-swapping},
  author={Shiohara, Kaede and Yang, Xingchao and Taketomi, Takafumi},
  booktitle={Proceedings of the IEEE/CVF international conference on computer vision},
  pages={7634--7644},
  year={2023}
}

@article{stylegan3,
  title={Alias-free generative adversarial networks},
  author={Karras, Tero and Aittala, Miika and Laine, Samuli and H{\"a}rk{\"o}nen, Erik and Hellsten, Janne and Lehtinen, Jaakko and Aila, Timo},
  journal={Advances in neural information processing systems},
  volume={34},
  pages={852--863},
  year={2021}
}

@inproceedings{arcface,
  title={Arcface: Additive angular margin loss for deep face recognition},
  author={Deng, Jiankang and Guo, Jia and Xue, Niannan and Zafeiriou, Stefanos},
  booktitle={Proceedings of the IEEE/CVF conference on computer vision and pattern recognition},
  pages={4690--4699},
  year={2019}
}

@inproceedings{lpips,
  title={The unreasonable effectiveness of deep features as a perceptual metric},
  author={Zhang, Richard and Isola, Phillip and Efros, Alexei A and Shechtman, Eli and Wang, Oliver},
  booktitle={Proceedings of the IEEE conference on computer vision and pattern recognition},
  pages={586--595},
  year={2018}
}

@inproceedings{uap,
  title={Universal adversarial perturbations},
  author={Moosavi-Dezfooli, Seyed-Mohsen and Fawzi, Alhussein and Fawzi, Omar and Frossard, Pascal},
  booktitle={Proceedings of the IEEE conference on computer vision and pattern recognition},
  pages={1765--1773},
  year={2017}
}

@article{mixup,
  title={mixup: Beyond empirical risk minimization},
  author={Zhang, Hongyi and Cisse, Moustapha and Dauphin, Yann N and Lopez-Paz, David},
  journal={arXiv preprint arXiv:1710.09412},
  year={2017}
}

@inproceedings{m_mixup,
  title={Manifold mixup: Better representations by interpolating hidden states},
  author={Verma, Vikas and Lamb, Alex and Beckham, Christopher and Najafi, Amir and Mitliagkas, Ioannis and Lopez-Paz, David and Bengio, Yoshua},
  booktitle={International conference on machine learning},
  pages={6438--6447},
  year={2019},
  organization={PMLR}
}

\appendix

\end{document}